\documentclass[10pt,twocolumn,letterpaper]{article}

\usepackage{cvpr}              % To produce the CAMERA-READY version
\usepackage[dvipsnames]{xcolor}

\definecolor{cvprblue}{rgb}{0.21,0.49,0.74}
\usepackage[pagebackref,breaklinks,colorlinks,citecolor=cvprblue]{hyperref}

\usepackage{algorithmic}
\usepackage{graphicx} 
\usepackage{acro} 
\usepackage{listings}
\usepackage{framed}
\usepackage{rotating}
\usepackage{makecell}
\usepackage{blindtext}
\usepackage{multirow} 
\usepackage{subcaption}  
\usepackage{caption}
\usepackage{pifont}
\usepackage{xcolor}
\usepackage{threeparttable}
\usepackage{amssymb}% http://ctan.org/pkg/amssymb
\usepackage{pifont}% http://ctan.org/pkg/pifont
\newcommand{\cmark}{\ding{51}}%
\newcommand{\xmark}{\ding{55}}%
\usepackage{amssymb}
\usepackage{siunitx}
\usepackage{multicol}
\usepackage{tikz} 
\usepackage{tikzpagenodes}

\DeclareAcronym{ai}{short=AI, long= Artificial Intelligence}
\DeclareAcronym{ann}{short=ANN, long= Artificial Neuron Network}
\DeclareAcronym{cnn}{short=CNN, long=Convolutional Neuron Network} 
\DeclareAcronym{dvs}{short=DVS, long=Dynamic Vision Sensor}  
\DeclareAcronym{if}{short=IAF, long= Integrate And Fire}
\DeclareAcronym{lstm}{short=LSTM, long= Long-Short-Term-Memory} 
\DeclareAcronym{mac}{short=MAC, long= Multiply and Accumulate} 
\DeclareAcronym{nir}{short=NIR, long=Near-Infra Red (Illumination)} 
\DeclareAcronym{nas}{short=NAS, long=Neural Architecture Search} 
\DeclareAcronym{snn}{short=SNN, long= Spiking Neuron Network} 
\DeclareAcronym{soc}{short=SoC, long=System-on-Chip} 
\DeclareAcronym{speck}{short=Speck, long= Synsense Speck}  
\DeclareAcronym{yolo}{short=YOLO, long=You Only Look Once}

\definecolor{codegreen}{rgb}{0,0.6,0}
\definecolor{codegray}{rgb}{0.5,0.5,0.5}
\definecolor{codepurple}{rgb}{0.58,0,0.82}
\definecolor{backcolour}{rgb}{0.99,0.99,0.99}

\lstdefinestyle{mystyle}{
    backgroundcolor=\color{backcolour},
    commentstyle=\scriptsize\color{codegreen},
    keywordstyle=\scriptsize\color{blue},
    numberstyle=\scriptsize\color{codegray},
    stringstyle=\ttfamily\scriptsize\color{codepurple},
    basicstyle=\ttfamily\scriptsize,
    breakatwhitespace=false,
    breaklines=true,
    captionpos=b,
    keepspaces=true,
    numbers=left,
    numbersep=5pt,
    showspaces=false,
    showstringspaces=false,
    showtabs=false,
    tabsize=2
}

\def\paperID{2} % *** Enter the Paper ID here
\def\confName{CVPR}
\def\confYear{2024}

\title{Retina : Low-Power Eye Tracking with Event Camera and Spiking Hardware}

\author{
    Pietro Bonazzi$^{1}$, Sizhen Bian$^{1}$ ,  Giovanni Lippolis$^{2}$, Yawei Li$^{1}$, Sadique Sheik$^{3}$ , Michele Magno$^{1}$ \\
    $^{1}$ETH Z\"urich, $^{2}$Inivation AG, $^{3}$Synsense AG \\ 
}

\begin{document}

\maketitle

\begin{abstract}
This paper introduces a neuromorphic dataset and methodology for eye tracking, harnessing event data captured streamed continuously by a \ac{dvs}. The framework integrates a directly trained \ac{snn} regression model and leverages a state-of-the-art low power edge neuromorphic processor - Speck. First, it introduces a representative event-based eye-tracking dataset, "Ini-30," which was collected with two glass-mounted \ac{dvs} cameras from thirty volunteers. Then, a \ac{snn} model, based on \ac{if} neurons, named  "Retina", is described , featuring only 64k parameters (6.63x fewer than 3ET) and achieving pupil tracking error of only 3.24 pixels in a 64x64 \ac{dvs} input. The continuous regression output is obtained by means of temporal convolution using a non-spiking 1D filter slided across the output spiking layer over time. 
Retina is evaluated on the neuromorphic processor, showing an end-to-end power between 2.89-4.8 $mW$ and a latency of 5.57-8.01 $ms$ dependent on the time  to slice the event-based video recording. The model is more precise than the latest event-based eye-tracking method, "3ET", on Ini-30, and shows comparable performance with significant model compression (35 times fewer MAC operations) in the synthetic dataset used in "3ET". 
We hope this work will open avenues for further investigation of close-loop neuromorphic solutions and true event-based training pursuing edge performance.
\end{abstract}

\begin{figure}[htbp] 
    \centering
    \captionsetup{type=figure}
    \includegraphics[trim={0cm, 0cm, 0cm, 1.2cm}, clip, width=\linewidth]{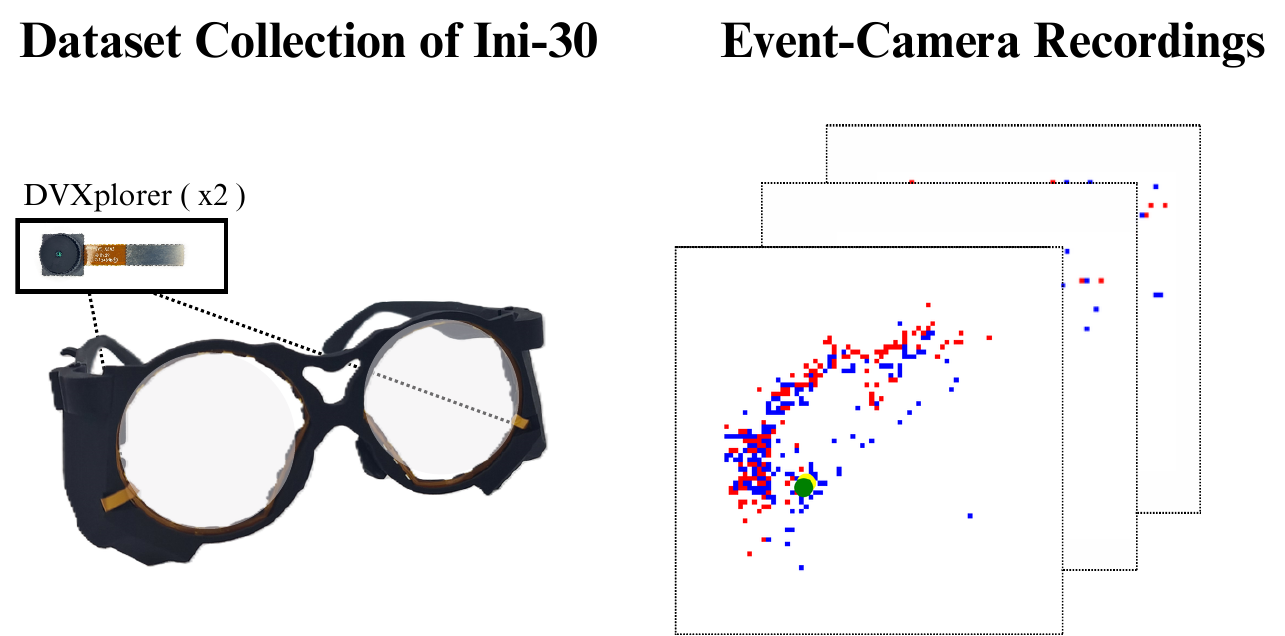}
    \captionof{figure}{A picture of the hardware for data collection (left) and an example of the video recordings (right) with ground truth (green) and prediction (yellow).}
    \label{fig:dataset_setup}
\end{figure}

\section*{Reproducibility}
\url{https://github.com/pbonazzi/retina} 

\section*{Acknowledgments}

This research was funded by Innosuisse (103.364 IP-ICT). We thank Wolfgang Böttcher, Adam Radomski and Nogay Küpelioğlu for the great help during the collection of the Ini-30 dataset and the Deployment on Speck. 
    
\section{Introduction}
\label{sec:intro}

% \begin{figure*}[t]
%     \centering
%     \begin{tabular}{cc}
%         \begin{subfigure}{0.75\linewidth}
%             \centering
%             \includegraphics[trim={1.5cm, 2.5cm, 0cm, 1.1cm}, clip, width=\linewidth]{images/thumbnail.pdf}
%             \caption{Architectural overview of the proposed method.}
%         \end{subfigure} &
%         \begin{subfigure}{0.2\linewidth}
%             \centering
%             \includegraphics[trim={0cm, 0cm, 0cm, 0cm}, clip, width=\linewidth]{images/scores.pdf}
%             \caption{Key results.}
%         \end{subfigure} \\
%     \end{tabular}
%     \caption{The processing pipeline of our method. Event-based efficient eye tracking is achieved by training a spiking neural network on our proposed dataset and deploying it on a neuromorphic \ac{soc}.}
% \end{figure*}

Neuromorphic systems, mimicking the neurobiological architectures of the human brain, have emerged as a promising paradigm for sensing and processing \cite{Bulzomi_2023_CVPR, Abreu_2023_CVPR, Schnider_2023_CVPR}. Event cameras and spike-based computation present distinctive advantages compared to traditional frame-based cameras and \ac{ann}, such as low power and decreased computing complexity. In the context of neuromorphic systems for eye tracking, various challenges confront the field. First, even the most recent methods either rely on frame-based input \cite{chen20233et, zhao2023ev} or focus on end-to-end tasks, such as gaze tracking learned from fixed screen-level coordinates \cite{angelopoulos2020event}. However, pupil tracking serves as a crucial initial phase in the gaze estimation pipeline. When mastered, it can facilitate applications to extend their capabilities beyond controlled environments. 
Second, state-of-the-art algorithms for event-based eye tracking are model-based and require subject-specific calibration \cite{angelopoulos2020event, zhao2023ev} and fitting at the moment of use, thus limiting their generalization to consumer application \cite{zhang2019evaluation}. 
Furthermore, eye-tracking algorithms and systems demand a nuanced equilibrium and trade-off between resolution, frame rate, latency and power consumption, as each pixel carries energy and bandwidth costs during the acquisition process.
One of the most recent papers on event-based eye tracking without subject-dependent calibration was authored by Chen et al. \cite{chen20233et}. Notably, in this work, they accumulated and normalized synthetically generated events to a 32-bit pixel resolution frame. This approach is inefficient as it does not leverage the asynchronous 1-bit nature of the event data. Finally, there is a need for end-to-end assessments of power consumption and latency, with the object of benchmarking the energy efficiency and real applicability of machine learning algorithms, that leverage event input and spike-based computation.

To address these challenges, we propose an eye-tracking model, dubbed "Retina", suitable for deployment on a neuromorphic asynchronous \ac{soc}, \ac{speck}. We leverage spike-based computation and real event input to offer energy-efficient eye tracking of the pupil from near-eye events.  In detail, our work introduces the following three key contributions: 

\begin{enumerate}

\item  \textbf{Event-based Eye Tracking Dataset, "Ini-30"}:
We introduce the first event-based eye-tracking dataset, named "Ini-30", recorded on a glass frame. Our dataset is collected with two event cameras and features variable recording lengths and event counts from 30 volunteers, providing an ideal benchmark for modeling the heterogeneity of event-based eye tracking in real-world scenarios. Our dataset brings new challenges, such as different event temporal evolution trends across different recordings. Thus, we slice the event data based on quantity of events instead of fixed timestamps. This method enables domain gap adaptation between different \ac{dvs}, solves the difference in temporal event growth between real recordings, and improves eye tracking precision.

\item \textbf{Event-based Eye Tracking Algorithm, "Retina"}:
Retina is a \ac{snn} structure based on \ac{if} neurons, followed by a non-spiking temporal weighted-sum filter for regression, which converts spikes to bounding box predictions. The filter allows \ac{if} neurons to learn temporal information without having to fall back to a voltage decay factor \cite{bos2023sub} or recurrent neurons, which are not supported in the hardware, \ac{speck}. To the best of our knowledge, Retina is the first eye tracking algorithm, suitable for deployment on neuromorphic hardware. Compared to 3ET \cite{chen20233et}, it shows superior precision (-20\% centroid error) and reduced computational complexity (-30x MAC).  

\item \textbf{Deployment on Neuromorphic Hardware}:
Finally, the deployment of our model on an "edge" neuromorphic \ac{soc}, \ac{speck}, provides for the first time power and latency results for network inference on chip and an end-to-end evaluations using the on-board \ac{dvs} camera. 

\end{enumerate}
\section{Related Work} 

%The exploration of eye-tracking could be traced back to the beginning of the twentieth century when an eye-movement tracking device was built during the reading process \cite{huey1908psychology}. Over one century of continuous research, especially triggered by emerging applications in fields like psychopathology, optometry, human-computer interaction, etc., eye-tracking techniques and systems have already been driven into real business applications, though still facing challenging requirements in accuracy, update rate, cost, portability, etc.\cite{holmqvist2023eye, plopski2022eye}. 
This section discusses the state-of-the-art explorations of eye-tracking based on a comprehensive literature survey. Depending on the utilized signal form (non-event or event input), we separate the existing eye-tracing solutions into two approaches:

\begin{table*}[t]
\centering
     \begin{threeparttable}

\caption{A summary of the related work in event-based pupil/gaze tracking}
\label{table:relatedwork}
\small
\begin{tabular}{ p{1.4cm} p{1.3cm} p{1.5cm} p{2.0cm}  p{2.0 cm} p{1.6cm}  p{1.8cm}  p{1.6cm} }
\toprule 
Year-Work & \ac{dvs} Camera  & Dataset  & Input & Algorithm  & Event Rate   & Energy  & Precision \\ 
 \midrule 
 
2020-\cite{angelopoulos2020event}  & DAVIS346b  & Customized  & Frames/Events  & Model-based   & $\geq$ 10 $kHz$ & N/A \tnote{a}  & 0.45$^{\circ}$–1.75$^{\circ}$ \\  \midrule 
2022-\cite{stoffregen2022event}  & Prophesee G3.1 & N/A & Events/LED Markers  & Glint Detection & 1 $kHz$ & $\approx$ \SI{35}{\milli\watt}\tnote{b} &  $<$ 0.5px \\  \midrule 
2023-\cite{zhao2023ev}  & DAVIS346 & EV-Eye & Frames/Events & Point2Edge  & $\leq$ 38.4 $kHz$  &  N/A & 1.2-7.7\si{px}  \\  \midrule 
2023-\cite{chen20233et} & N/A & Synthetic & Event-frames & ConvLSTM & 95 $kHz$  & N/A & N/A  \\  \midrule 
\textbf{Ours} & DVXplorer & Ini-30 & Events & \ac{snn} & $\leq$ 5$kHz$ \tnote{c} & 2.89\tnote{d}-4.8$mW$\tnote{d}  & 3-8px \\ 

\bottomrule
\end{tabular}
\begin{tablenotes}
\setlength{\columnsep}{0.8cm}
\setlength{\multicolsep}{0cm}
  \begin{multicols}{2}

            \item[a] Not Available.
            \item[b] Sensor power only.
            \item[c] Depending on the \ac{dvs} and time window lenght.    
            \item[d] End-to-end power. 

  \end{multicols}
\end{tablenotes}
\end{threeparttable}

\end{table*}

\subsection{Non-Event-Based Eye Tracking}

Conventional non-event-based eye-tracking includes model-based and appearance-based methods. The former one either tracks specular glint reflections for corneal curvature center detection \cite{turetkin2022real, hosp2020remoteeye}, or extracts salient geometrical features of the eye from frames and tracks the pupil with an optimized fitting method based on a physics eye model \cite{wang2017real, khan2019gaze}. Such an approach commonly supplies impressive tracking accuracy, achieving sub-degree tracking error \cite{mestre2018robust}, while stunted by the deployment complexity, such as ambient light conditions, image resolution, and calibration requirement. The latter one typically applies a trained machine-learning model to the raw eye images for tracking \cite{kellnhofer2019gaze360, zhang2019evaluation, palmero2018recurrent, tinytracker2023}. This approach gives an end-to-end, deployable eye-tracking solution, which is heavily limited by the frame rate of the camera, with resulting tracking rate that can reach a maximum of 300 Hz.  Differently from this line of work, the event-based eye tracking can reach beyond $kHz$ \cite{angelopoulos2020event} of update rate and provide energy-efficient observations \cite{survey2019events}.

\subsection{Event-Based Eye Tracking}

Benefiting from the sparse event stream and high dynamic range of event cameras, event-based eye-tracking has emerged as a groundbreaking solution enabling beyond-$kHz$ and low-power consumption eye-tracking, as listed in \cref{table:relatedwork}. The first event-based eye-tracking work was published in 2020 \cite{angelopoulos2020event}, in which the authors proposed a hybrid frame-event-based near-eye gaze tracking system offering update rates beyond 10 $kHz$ with an accuracy comparable to commercial tracker. The algorithm is based on a parametric pupil model and utilizes the event to update the model with a pupil-fitting method. This work is sensitive to sensor noises, as choosing the useful event that can be used to update the model is challenging.  Following this work, researchers have published another three event-based pupil/gaze-tracking papers in the past three years. 
%Ryan et al. \cite{ryan2021real} presented a method to detect and track faces and eyes for driver monitoring simultaneously using event frames formed voxel grid from an event camera. 
Stoffregen et al. \cite{stoffregen2022event} described the first fully event and model-based glint tracker, which is robust to background disturbances and has a sampling rate of 1 $kHz$ with an estimated power of 35 $mW$ (sensing components only). Here the author used coded differential lighting to enhance the glint detection with an event camera. Similar to \cite{angelopoulos2020event}, in \cite{zhao2023ev}, the authors presented a hybrid eye-tracking method that leverages both the near-eye grayscale images and event data for robust and high-frequency eye tracking. The proposed matching-based pupil tracking method gave a pixel error of only 1.2 px with a peak tracking frequency of up to 38.4 $kHz$. In \cite{chen20233et}, the authors proposed a sparse change-based convolutional \ac{lstm} model for event-based eye tracking, which reduces arithmetic operations by approximately 4.7×, compared to a standard convolutional \ac{lstm}, without losing accuracy when tested on a synthetic event dataset. \newline
There are several limitations to existing methods. First, they rely on frames, either directly from the sensor output  \cite{angelopoulos2020event, zhao2023ev}, which decreases the power efficiency of the system, or accumulate the events into frames and process the frames with deep neural network \cite{chen20233et}, which sacrifices the temporal resolution, introduces latency and increase the memory footprint.  
Second, with a purely event-based solution, auxiliary devices are used to enhance specular events \cite{stoffregen2022event}. Besides that, none of the existing works carried out a real deployment of their proposals, especially on a neuromorphic platform that perfectly matches the sparsity of the event stream. Thus, the system-level performance in the energy and latency that an event solution can bring is still unclear. 
In contrast, our work presents a \ac{snn} supported by an end-to-end neuromorphic edge system leveraging pure events stream from a \ac{dvs} camera. 
\section{Dataset} 
\label{sec:dataset}
Several prominent eye-tracking and gaze estimation datasets have contributed significantly to the advancement of this field using frame cameras \cite{krafka2016eye, zhang2018agil}. 
To the best of our knowledge, the only available event-based datasets have been presented by Angelopoulos et al. \cite{angelopoulos2020event} for \textit{gaze tracking} and in Zhao et al. \cite{zhao2023ev} for \textit{gaze and eye tracking}. As a matter of fact, the state-of-the-art methods, i.e. 3ET \cite{chen20233et}, had to generate synthetic event-based dataset to develop eye tracking algorithms. In this paper, we introduce a representative event-based \textit{eye tracking} dataset, dubbed "Ini-30", which, for the first time, was collected with two event cameras (one per eye) mounted on a glass-frame without fixing the head of the user on a head set in a controlled environment.  

\subsection{Dataset Collection}

The Ini-30 dataset is collected with two event cameras mounted on a glass frame. Each DVXplorer sensor (640 × 480 pixels) is attached on the side of the frame. The power supply was provided via a 2 meter cable connected from the cameras to a computer, which provided enough freedom of movement. Differently from \cite{angelopoulos2020event, zhao2023ev}, the participants were not instructed to follow a dot on a screen, but rather encouraged to look around to collect natural eye movements. As shown in \cref{fig:dataset_setup}, the event cameras were securely screwed on a 3D-printed case attached to the side of the glass frame. 

The data was annotated based on accumulated linearly decayed events by defining the pixel intensity as function of the linear accumulation of previous pixel intensity. Next we labeled the position of the pupil in the \ac{dvs}'s array manually, using an assistive labeling tool.  We discarded the first 20ms of events to ensure the eye was visible and annotations met the level of image-based annotators.
The number of labels per recording was intentionally variable, spanning from 475 to 1'848 with a time per label ranging from 20.0 to 235.77 milliseconds depending on the overall duration of the sample.

This setup allows for unconstrained head movements, enables to capture event data from eye movement in a "in-the-wild" setting and allows the generation of a representative, unique, diverse and challenging dataset. In the next section, we characterize the dataset and compared further with event-based datasets for gaze tracking.

\subsection{Dataset Comparaison}

In this section, we present a comparative analysis of the dataset presented in \cite{angelopoulos2020event, zhao2023ev} with our proposed Ini-30 dataset. We summarize the key points in \cref{tab:ange_vs_ours}. 
To the best of our knowledge, Ini-30 is the first eye-tracking event-based dataset with pupil location labelled on the sensor. The other event-based datasets available \cite{angelopoulos2020event, zhao2023ev} consist of point coordinates on a display, instead of pupil location coordinates on the \ac{dvs}'s array. Labeling pupil locations provides superior precision and granularity for understanding gaze behavior compared to screen coordinates. 

\begin{table}[htbp]
\small
    \caption{Comparison between our proposed dataset (Ini-30) with the one proposed in \cite{angelopoulos2020event} and \cite{zhao2023ev}.\strut}
    \label{tab:ange_vs_ours}
    \begin{tabular}{l|p{0.3\linewidth}|p{0.25\linewidth}} \toprule
        Aspect &  \cite{angelopoulos2020event} \cite{zhao2023ev} & Ini-30 (Ours)  \\ \midrule  
            Resolution & 346×260px  & 640×480px \\  
            Glass Frame & \textcolor{red}{\xmark} & \textcolor{green}{\cmark} \\
            Pupil Label & \textcolor{red}{\xmark} & \textcolor{green}{\cmark}\\ 
            Variability* & Low & High \\ 
        \bottomrule
    \end{tabular} 
    \vspace{1ex}
    
    {\raggedright * Variability refers to the difference in duration of each recordings and the event temporal distribution. \par}
\end{table}

%\begin{figure}[htbp] 
 %   \centering
  %  \captionsetup{type=figure}
   % \includegraphics[trim={0cm, 0cm, 0cm, 0cm}, clip, width=0.7\linewidth]{images/dataset/temp_evol.pdf}
    %\captionof{figure}{Event temporal evolution per recording in Ini-30.}
    %\label{fig:temp_evo_ours}
%\end{figure}

%\begin{table}[h]
%\small
%\centering
 %   \caption{Event temporal evolution per recording in Ini-30 (left) and \cite{angelopoulos2020event, zhao2023ev} (right).}
  %  \label{fig:temp_evo_ours}
   % \begin{tabular}{c c} 
    %    \raisebox{-.5\height}{\begin{subfigure}{0.45\linewidth}
     %        \includegraphics[trim={1cm, 1cm, 1cm, 1cm}, clip,width=\linewidth]{images/ini30_dataset.png}
      %  \end{subfigure}} &
       % \raisebox{-.5\height}{\begin{subfigure}{0.45\linewidth}
        %     \includegraphics[trim={1cm, 1cm, 1cm, 1cm}, clip,width=\linewidth]{images/berkley_0.png}
 %       \end{subfigure}} \\ 
%    \end{tabular} 
%\end{table} 

In addition, \cite{angelopoulos2020event, zhao2023ev} employs a lower resolution sensor, DAVIS346b (346 × 260 px), whereas our Ini-30 dataset incorporates DVXplorer sensor, 640 × 480 px. Both datasets are collected using \ac{nir} illumination technology, to highlights the events surrounding the pupil. Additionally,  \cite{angelopoulos2020event, zhao2023ev} captures subjects with fixed head positions, while Ini-30 is designed for use cases with unconstrained head movements with cameras mounted directly on glass frames. Because of this, Ini-30 covers a broader range of situations. For example, temporally,  \cite{angelopoulos2020event} maintains fixed recording lengths ($30s$) and linearly growing event counts, while Ini-30 showcases richer variability, \cref{tab:ts}, with recording durations spanning from 14.64 to 193.8 seconds.

\cref{tab:ts} provides key statistics for Ini-30, in the context of event cameras updates rates. The sampling time step ranges from 61 to 346 microseconds, and the event count per timestep spans from 3 to 5'000 events.

\begin{table}[htbp]
\small
    \caption{Statistics for sampling times (Ts [${\mu s}$] ) and number of events per timestamps in Ini-30.}
    \label{tab:ts}
    \begin{center}
        \begin{tabular}{l|c|c|c|c|c} \toprule 
             Name& Median & Mean & Std  &  Min &  Max\\  \midrule  
            Sampling Time & 200 & 200 &  14 & 61 & 346 \\
            Events / Ts & 94 & 175 &  299 & 3 & 4'799 \\ \bottomrule
        \end{tabular}
    \end{center} 
\end{table}

Overall, our dataset encompasses a diverse range of recordings (30), with labels per recording spanning from 475 to 1'848, with total event data varying from 4.8 million to 24.2 million, and time per label ranging from 20.0 to 235.77 milliseconds. This information helped informed data preparation strategies in our models which consider a temporal dimension.  
\section{Methodology} 
\label{sec:methodology}

We propose a low latency and lightweight architecture and learning rule for an eye-tracking algorithm based on event data.  Our network is a single \ac{snn} featuring spiking spatial convolutions and fusible batch normalization layers. The spiking outputs are converted to continuous values by means of a 1D convolution layer with fixed weights. An overview of our network configurations can be found in Tab \ref{tab:network_memory_core}, while other implementation details are described in the next subsections.

\subsection{Data Preparation} 
\label{sec:mec-preprocessing}

Since the algorithm is deployed on the neuromorphic \ac{soc} \ac{speck}, which has two-channel support for a 64x64 \ac{dvs}'s resolution, we prepared the dataset to bridge this domain gap. 
First, we transformed the rectangular resolution of the original data to a squared array of 512x512, by shifting the y-axis to 16 pixels and discarding 128 X-coordinates in correspondence to the spatial location where fewer events are present and no label appeared ($x<96$ and $x>608$).

\begin{figure}[b]
    \centering
    \includegraphics[trim={1.5cm, 4cm, 12.5cm, 2.5cm}, clip, width=\linewidth]{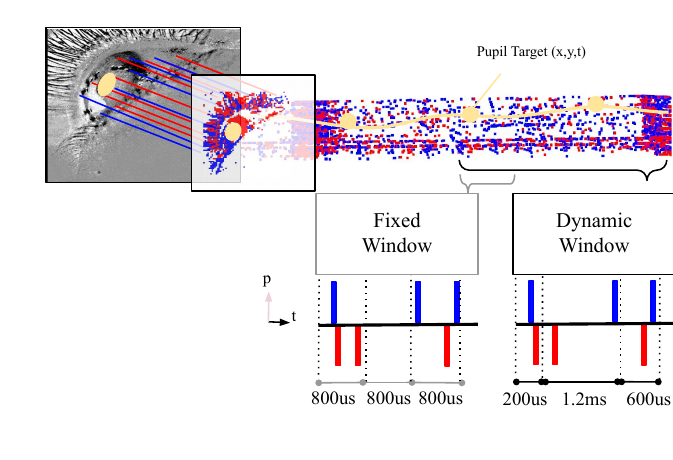} 
    \caption{An example illustrating the different techniques for slicing events video recordings (red, blue) in time: A) dt = $800us$, B) events count = $2$.}
    \label{fig:distribution-events-fixed-window}
\end{figure}

Next, we applied sum pooling to reach the \ac{speck} compatible resolution and proceeded to dynamically slice the event temporally. Every video was sliced, from a point in time corresponding to a pupil label timestamp, until we obtained a desired pixel activations. This results in better input data for convolution layers, (\cref{fig:event-slicing}), \cref{fig:distribution-events-fixed-window} shows an illustrated example of the working principles of the two techniques. 
%, w.r.t fixed time windows and redistributes the load, in terms of event quantities on the neuromorphic chip \cref{fig:distribution-events-fixed-window}. 
% \cref{fig:distribution-events-fixed-window} shows an illustrated example of the working principles of the two techniques. 
In case multiple events from the same pixels are found during this slicing step, we keep the event polarity with the highest number of events and then clip the event frame back to 0 and 1, ensuring only one channel is active at every timebin. Finally, since in our dataset, the sampling time of events has a median of 200 ${\mu s}$, \cref{tab:ts}, while pupil labels have a frequency of around 30 ${ms}$, we weight-interpolated new labels at bin time using the two closest labels in the source recordings. We trained our network with 64 timebins, shuffling each slice of recording only during training.

% \begin{figure}
%     \centering
%     \includegraphics[width=\linewidth]{sec/variability.pdf} 
%     \caption{A plot showing the distribution of events per timestamp
% on the validation set using fixed time window of 3ms.}
%     \label{fig:distribution-events-fixed-window}
% \end{figure}

% \begin{lstlisting}[language=Python, caption=The code to dynamically slice event windows in time, label=python_code]
% # data preparation
% import numpy as np 
% XYP=last_events_before_pupil_label
% end_index = len(XYP) - 1
% # event bins
% tcwh=np.zeros(64,2,64,64))
% for i in reversed(range(64)): 
%  # events for time bin (i)
%  x,y=last_unique_evs(XYP[:end_index,:2],N)
%  start_index=end_index-len(x)
%  p=XYP[start_index:end_index]
%  # add events for i
%  np.add.at(tcwh[i,0],(x[p==0],y[p==0]),1)
%  np.add.at(tcwh[i,1],(x[p==1],y[p==1]),1) 
%  # recover 1-bit event channel
%  tcwh[i,0][tcwh[i,1]>=tcwh[i,0]]=0 
%  tcwh[i,1][tcwh[i,1]<tcwh[i,0]]=0
%  tcwh[i]=tcwh[i].clip(0, 1) 
%  # move index
%  end_index = start_index   
% \end{lstlisting}

% \begin{table}[htbp]    
%     \caption{A graph illustrating the different techniques for slicing events (red, blue) in time: A) dt = $800us$, B) evs = $2$.}
%     \label{fig:fixed-window}
%     \begin{tabular}{c} 
%         \begin{subfigure}{\linewidth}
%             \includegraphics[trim={0cm, 6cm, 14cm, 0cm}, clip, width=\linewidth]{images/timeslicing.pdf} 
%             \label{fig:img2}
%         \end{subfigure}
%     \end{tabular}   
% \end{table}

\begin{table}[htbp]  
    \caption{An example of the qualitative improvements of  slicing video recording with a dynamic time window (1) compared to fixed time (2).}
    \label{fig:event-slicing}
    \begin{tabular}{c|c|c|c} 
    \toprule
        Method & $t$ & $t_{i+1}$ & $t_{i+2}$ \\  \midrule
        (1) &
        \raisebox{-.5\height}{
        \begin{subfigure}{0.2\linewidth}
              \includegraphics[ width=\linewidth]{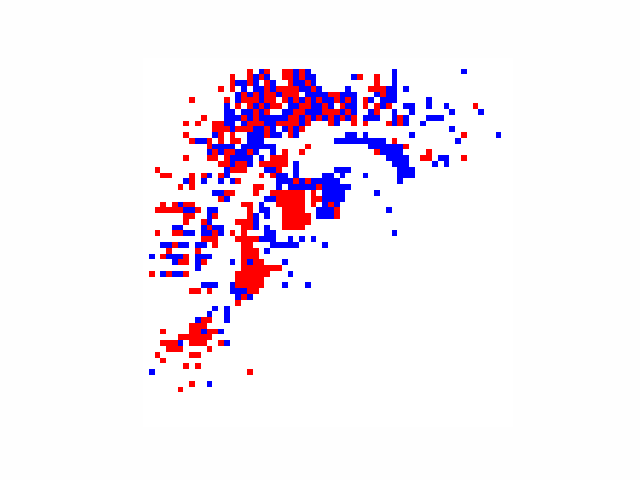}
        \end{subfigure}}&
        \raisebox{-.5\height}{
        \begin{subfigure}{0.2\linewidth}
             \includegraphics[ width=\linewidth]{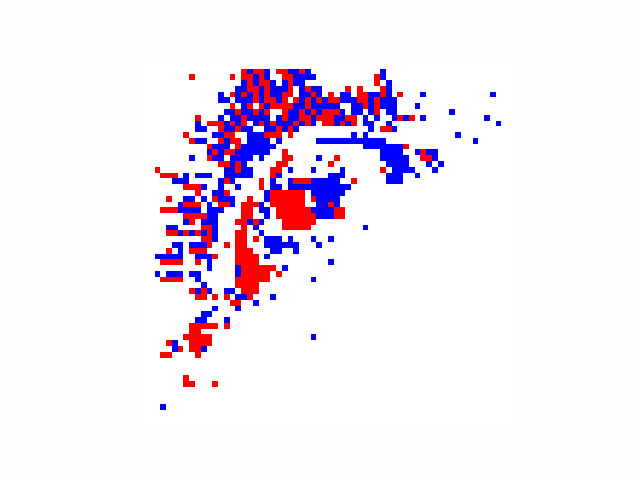}
        \end{subfigure}}&
        \raisebox{-.5\height}{
        \begin{subfigure}{0.2\linewidth}
              \includegraphics[ width=\linewidth]{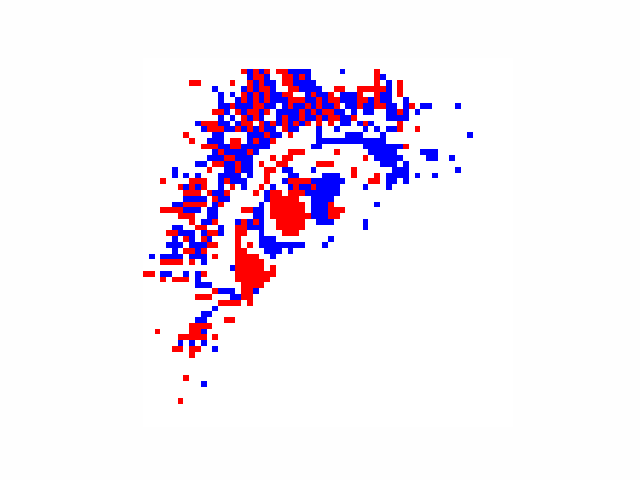}
        \end{subfigure}}\\ \midrule
        (2) &
        \raisebox{-.5\height}{
        \begin{subfigure}{0.2\linewidth}
             \includegraphics[ width=\linewidth]{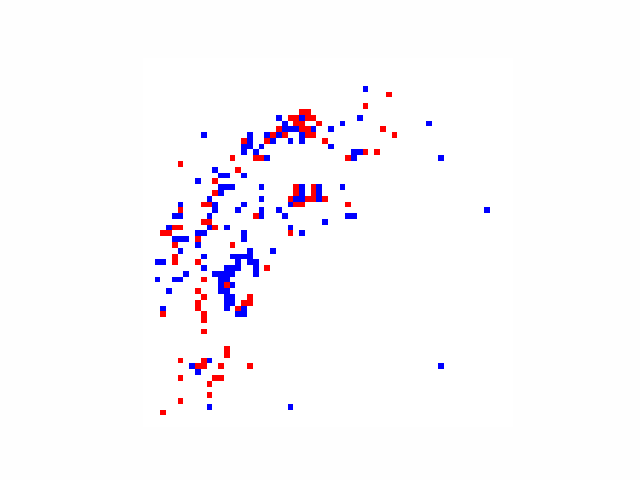}
        \end{subfigure}}&
        \raisebox{-.5\height}{
        \begin{subfigure}{0.2\linewidth}
              \includegraphics[ width=\linewidth]{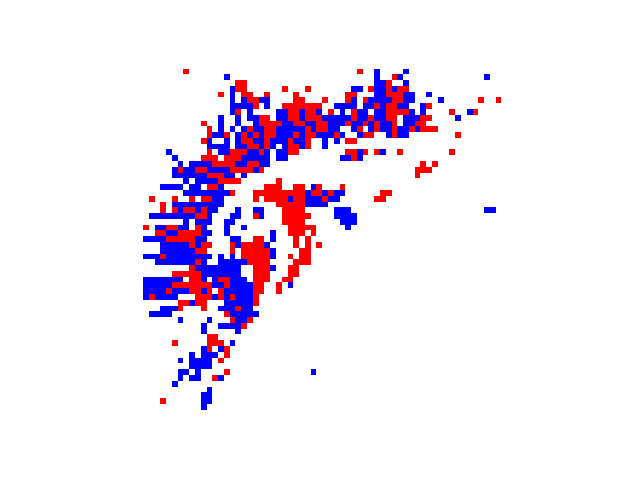}
        \end{subfigure}}&
        \raisebox{-.5\height}{
        \begin{subfigure}{0.2\linewidth}
           \includegraphics[ width=\linewidth]{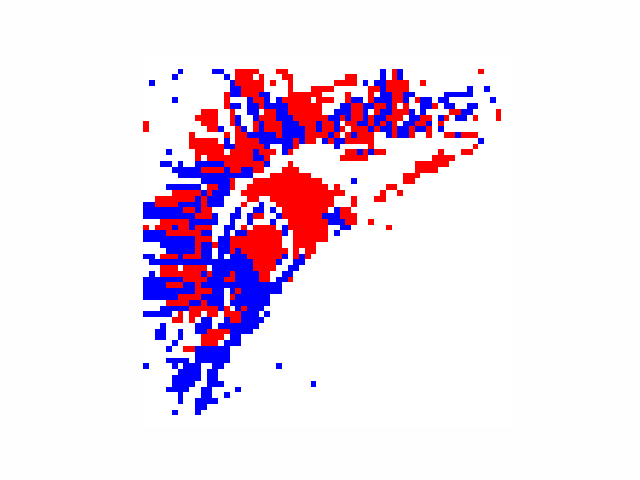}
        \end{subfigure}}\\  \bottomrule
    \end{tabular}  
\end{table}

\subsection{Network Architecture}

After the data preparation step, events are processed sequentially at fixed timestamps by kernel of the convolution, batch normalization layers and \ac{if} neurons in our model. We designed each layer of our network to fit the memory limitation of the available cores in the targeted platform, \ac{speck}. 

% \begin{table}[htbp]  
%     \caption{The memory limits of each core in  \ac{speck} are reported in kibibyte (1 Ki = 1024 bytes).}
%     \label{tab:cores}
%     \small
%     \begin{tabular}{c|c|c}
%         \toprule
%         Core ID & $K_{MT}$ (Ki) & $N_{M}$ (Ki) \\
%         \midrule
%         0 & 16  & 64  \\
%         1 & 16  & 64  \\
%         2 & 16  & 64  \\
%         3 & 32  & 32  \\
%         4 & 32  & 32  \\
%         5 & 64  & 16  \\
%         6 & 64  & 16  \\
%         7 & 16 & 16  \\
%         8 & 16 & 16  \\
%         \bottomrule
%     \end{tabular}  
% \end{table} 

We computed the total kernel memory available ${K_{MT}}$ for each layer following the basic formula:
\begin{equation}
{K_{MT} = c × 2^{log_{2}^{k_{x}k_{y}} + log_{2}^{f}}}
\end{equation}
Where $c$ is the input channel of $M$, $f$ is the output channel number, ${k_x}$ and ${k_y}$ are the kernel size. The necessary neuron memory $N_{M}$ entries are computed by solving the following formulas: 
\begin{equation}
{N_{M} = f × f_{x} × f_{y}}
\end{equation}
Where  $f$ is the output channel number and $f_x$ and $f_y$ depend on the input feature map size $c_{x}$, $c_{y}$, stride $s_{x}$,$ s_{y}$, and padding $p_{x}$,$ p_{y}$, following the relationships:
\begin{equation}
{f_x = \frac{c_{x}-k_{x}+2p_{x}}{s_{x}} + 1}, 
{f_y = \frac{c_{y}-k_{y}+2p_{y}}{s_{y}} + 1}
\end{equation}

In \cref{tab:network_memory_core}, we present an overview of the network. The events are processed by Layer ID "1" and sequentially transmitted to the following layers. Every convolution operation is followed by a spiking neuron with a spiking threshold of 1 and a minimum voltage membrane of -1. Given that spike generation is non-differentiable, we use a surrogate gradient \cite{surrogate_2016} periodic exponential function \cite{weidel2021wavesense}. 

\begin{table*}[htpb] 
\centering
\caption{The network configuration, memory footprint and core compatibility on \ac{speck} for each layer.}
\small
\label{tab:network_memory_core}
    \begin{tabular}{c|c|c|c|c|c|c|c|c}
        \toprule
        Layer ID & SNN & $c_{in}$ & $c_{out}$ & $k_{x} * k_{y}$ & $s_{x} * s_{y}$ & $N_{M}$ & $K_{MT}$  & Cores ID\\
        \midrule
        % \multirow{2}{*}{0} & Conv* & 2 & 2 & $1 \times 1$ & $1 \times 1$ & 0.00 KiB & 8.51 KiB &  \multirow{2}{*}{all} \\
        % &\ac{if} (T=8)&  & && & & & \\
        % \hline
        \multirow{3}{*}{1} & BatchConv & 2 & 16 & $5 \times 5$ & $2 \times 2$ & \multirow{3}{*}{ 0.78 KiB} & \multirow{3}{*}{15.02 KiB} &  \multirow{3}{*}{all} \\
         & \ac{if}& & & & && & \\
         & Pool & && $2 \times 2$ & $1 \times 1$ & && \\
        \midrule
        \multirow{3}{*}{2}  & BatchConv &  16 & 64 & $3 \times 3$ & $1 \times 1$ & \multirow{3}{*}{9.00 KiB} & \multirow{3}{*}{64.00 KiB}  &  \multirow{3}{*}{0, 1, 2}\\
         &\ac{if}&  & && & & & \\
         & Pool & & & $2 \times 2$ & $1 \times 1$ & && \\
        \midrule
        \multirow{3}{*}{3}  & BatchConv &  64 & 16 & $3 \times 3$ & $1 \times 1$ &  \multirow{3}{*}{9.00 KiB} & \multirow{3}{*}{4.00 KiB}  &  \multirow{3}{*}{all}\\
         &\ac{if}&  & & && & & \\
         & Pool & & & $2 \times 2$ & $1 \times 1$ & && \\
        \midrule
        \multirow{2}{*}{4}  & BatchConv &  16 & 16 & $3 \times 3$ & $1 \times 1$ & \multirow{2}{*}{2.25 KiB} & \multirow{2}{*}{1.00 KiB} &  \multirow{2}{*}{all} \\
         &\ac{if}&  & & && & & \\
        \midrule
        \multirow{2}{*}{5}  & BatchConv &  16 & 8 & $3 \times 3$ & $1 \times 1$ & \multirow{2}{*}{1.12 KiB} & \multirow{2}{*}{0.50 KiB} &   \multirow{2}{*}{all}\\
         &\ac{if}&  & & & & && \\
        \midrule
        \multirow{2}{*}{6}  & BatchConv &  8 & 16 & $3 \times 3$ & $1 \times 1$ & \multirow{2}{*}{1.12 KiB} & \multirow{2}{*}{1.00 KiB} &   \multirow{2}{*}{all} \\
         &\ac{if}&  & & & && & \\
        \midrule
        \multirow{2}{*}{7} & BatchConv & 144 & 128 & $1 \times 1$ & $1 \times 1$& \multirow{2}{*}{18.00 KiB} & \multirow{2}{*}{1.12 KiB} &  \multirow{2}{*}{3, 4, 5, 6}\\
         & \ac{if}& & & && & & \\
        \midrule
        \multirow{2}{*}{8} & BatchConv & 128 & 160 & $1 \times 1$ & $1 \times 1$&  \multirow{2}{*}{34.37 KiB} & \multirow{2}{*}{2.42 Ki} & 5, 6 \\
         & \ac{if}& & & && & & \\
        \bottomrule
    \end{tabular} 
\end{table*}

The spikes generated by the final layer are converted to continuous values using a 1-dimensional non-spiking temporal weighted-sum filter with fixed weights, discussed in \cref{subsec:ssmoothing}. 
Next, akin to other grid-based methodologies such as those presented by Redmon et al. \cite{redmon2016look}, we modify the output layer to consist of $4\times4$ cells, each containing two 5-dimensional vectors representing the top right and bottom left coordinates of the bounding box, along with a confidence score. To refine the bounding box localization, we employ a post-processing step using Non-Maximum Suppression (NMS) \cite{NMS2019}, which helps eliminate redundant detections by retaining only the highest-scoring bounding boxes while discarding overlapping alternatives. In section \cref{subsec:loss}, we present how we compare the prediction to a generated target bounding box synthetically generated around the original 1-pixel label by expanding it to 2 pixels in each direction. 

\subsection{Neuron Model} 
\label{subsec:neuron-model}
The \ac{if} neuron model, operating after every convolution layer, is characterized by a straightforward mathematical formulation. It involves the integration of incoming synaptic inputs (convolution operations) and the generation of a spike once a certain membrane potential threshold is reached. The dynamics of the \ac{if} neuron are described by the following differential equation:
\begin{equation}
\tau_m \frac{dV}{dt} = -V(t) + R_m I_{\text{syn}}(t)
\end{equation}

where $V(t)$ is the membrane potential at time $t$, $\tau_m$ is the membrane time constant, $R_m$ is the membrane resistance, and $I_{\text{syn}}(t)$ represents the synaptic input current. The neuron fires when $V(t)$ surpasses a predefined threshold $V(th)=1$, at which point the membrane potential is reset to a resting value $V(reset)=0$. One of the key features of the \ac{if} neuron model is its integration mechanism for incoming synaptic inputs. The synaptic input $Isyn(t)$ is often modeled as a sum of weighted contributions from different synapses:

\begin{equation}
I_{\text{syn}}(t) = \sum_{j} w_j \cdot I_j(t - t_j)
\end{equation}

where $w_j$ represents the synaptic weight, $I_j(t - t_j)$ is the synaptic input spike train arriving at time $t$ from synapse $j$ with a spike at $t_j$.

\subsection{Temporal Weighted-Sum Filter} 
\label{subsec:ssmoothing}

The implemented temporal weighted-sum filter can be described as follows:

\begin{equation}
y(t) = \sum_{i=1}^{N} w_i \cdot x(t-i)
\end{equation}

where \(y(t)\) represents the filtered output at time \(t\), \(N\) is the length of the convolutional kernel, \(w_i\) denotes the filter weight at position \(i\) in the kernel, and \(x(t-i)\) is the input value at time \(t-i\).

The filter weights (\(w_i\)) are determined by a 'synaptic kernel' S(t) and 'membrane kernel' M(t), which in turn are computed based on a membrane constant (\(\tau_{\text{mem}}\)) and a synaptic constant (\(\tau_{\text{syn}}\)):

\begin{equation}
S(t) = \exp\left(-\frac{t}{\tau_{\text{{syn}}}}\right), 
M(t) = \exp\left(-\frac{t}{\tau_{\text{{mem}}}}\right)
\end{equation}

The membrane kernel initializes the weights of a 1D convolution applied to the synaptic kernel. The weights \(w_i\) of the temporal weighted-sum filter \(w_i\) are initialized as a result of the synaptic and membrane kernel convolution.

\subsection{Loss Function} 
\label{subsec:loss}
Our loss function  $\mathcal{L}$ is a combination of several components: a box loss $\mathcal{L}_{box}$, a confidence loss $\mathcal{L}_{conf}$, a synaptic loss  $\mathcal{L}_{syn}$ . 

The box loss $\mathcal{L}_{box}$ measures how well the model can localize the pupil within the image, by minimizing the mean squared error distance between the predicted \(p_i\) and target \(t_i\) bounding boxes, represented as:

\begin{equation}
  \mathcal{L}_{\text{box}} = \sum_{i=1}^{N} (p_i - t_i)^2
\end{equation}

The confidence loss  $\mathcal{L}_{conf}$ measures how confident the model is about its prediction, by penalizing low confidence scores for correct predictions and high confidence scores for incorrect predictions. This component is calculated as a mean squared error distance between the predicted and target confidence scores (\(c_i\) and \(g_i\)):

\begin{equation}
  \mathcal{L}_{\text{conf}} = \sum_{i=1}^{N}(c_i - g_i)^2
\end{equation}

The synaptic loss $\mathcal{L}_{syn}$ is the first of our regularization terms and it ensures that the number of multiply-accumulate operations performed by the neurons in the network at each layer is within a range the neuromorphic \ac{soc} \ac{speck} can handle (1e6).  $\mathcal{L}_{syn}$ is formulated as the squared difference between the target synaptic operations  with each layer synaptic operations, normalized by the square of the target synaptic operations. The total loss $\mathcal{L}$ is a weighted sum of these components. To summarize, the  $\mathcal{L}_{box}$and  $\mathcal{L}_{conf}$ are tasks losses which are used to detect the pupil in the event array. The  $\mathcal{L}_{syn}$ is a regularization component in order to deploy the network on the neuromorphic device. 
\section{Experiments}

\subsection{Setup} 

Our precision results are based on the centroid error in pixels extracted from the predicted bounding box. We evaluated the efficiency of our algorithm by measuring the power (P) [mW] consumption, energy (E) [mJ] consumption and the latency (L) [ms] on the neuromorphic \ac{soc}. In addition, we evaluated the network complexity with the number of parameters and \ac{mac} operations.  Regarding our implementation, we trained Retina's convolution layer ${M}$ with 8-bit weight parameters (kernel memory $K_{MT}$) and a 16-bit spiking neuron states (neuron memory $N_{M}$).  Batch normalization layers are fused to the convolution blocks in inference. The models are trained for 576 iterations using the ADAM optimizer and a step learning rate scheduler with a gamma of 0.8. Furthermore, we reset the states of the neuron at every iteration. The weights of the losses are $\lambda_{box}=7.5$, $\lambda_{conf}=1.5$ and 1e-7 for $\lambda_{syn}$. Finally,  we train our models with a batch size of 16, a sequence length of 64 , and an initial learning rate of 1e-3. The training takes 1 hour on a single NVIDIA GeForce RTX 4090.

\subsection{Ablation Studies} 

This section examines the two main components of our methodology: the dynamic event windows, the temporal weighted-sum filter as well as the loss function. The ablation studies are performed on the Ini-30 Dataset and using the best models. 

\subsubsection{Event-Based Video Recording Slicing}
\label{sec:exp-ablation-model}

\paragraph{Precision Results:} In \cref{table:slicing-methods}, we evaluate the effects of fixed time windows ($dt$) on the precision of pupil localisation. The evaluation considers the median, the minimum and maximum number of events/time-window per bin, see \cref{tab:ts}. The comparison is carried out considering the average time window. The dynamic event time window consistently outperform the fixed time window on the validation set.

\begin{table}[h]
\centering
\caption{The performance of different event slicing methods.}
\label{table:slicing-methods}
\small
\begin{tabular}{c|c|c|c}
\toprule
Method & Event Count & Time Window & Error (px)  $\downarrow$ \\ 
 \midrule 
\multirow{2}{*}{Fixed}   
& 198 & \multirow{2}{*}{$1.1ms$} & \multirow{2}{*}{4.40 ($\pm$ 2.69)}\\  
&  [$52,401$] &  &  \\ \midrule

\multirow{2}{*}{Dynamic}  & \multirow{2}{*}{100} & $1.1ms$ & \multirow{2}{*}{\textbf{3.54 ($\pm$ 1.43)}}\\  
&   & [$0.41,3.1ms$] &  \\ \midrule \midrule

\multirow{2}{*}{Fixed}  & 242 & \multirow{2}{*}{$1.6ms$} & \multirow{2}{*}{5.18 ($\pm$ 1.37)}\\  
&  [$69,481$] &  &  \\ \midrule

\multirow{2}{*}{Dynamic} & \multirow{2}{*}{150} & $1.6ms$ & \multirow{2}{*}{\textbf{3.49 ($\pm$ 1.18)}}\\  
&   & [$0.63,4.0ms$] &  \\   \midrule \midrule

\multirow{2}{*}{Fixed}  & 277 & \multirow{2}{*}{$2.1ms$} & \multirow{2}{*}{3.39 ($\pm$ 1.02)}\\  
&  [$84,541$] &  &  \\  \midrule

\multirow{2}{*}{Dynamic} & \multirow{2}{*}{200} & $2.1ms$ & \multirow{2}{*}{\textbf{3.46 ($\pm$ 1.36)}}\\  
&   & [$0.86,4.6ms$] &  \\  \midrule \midrule

\multirow{2}{*}{Fixed}  & 331 & \multirow{2}{*}{$3.0ms$} & \multirow{2}{*}{3.71 ($\pm$ 1.40)}\\  
&  [$110,630$] &  &  \\ \midrule

\multirow{2}{*}{Dynamic}& \multirow{2}{*}{300} & $3.0ms$ & \multirow{2}{*}{\textbf{3.24 ($\pm$ 0.79)}}\\  
&   & [$1.10,8.4ms$] &  \\  
\bottomrule
\end{tabular}
\end{table}  

\begin{figure}[h]
    \centering
    \begin{tabular}{c c}
        \begin{subfigure}{0.5\linewidth}
            \centering
            \includegraphics[trim={0cm, 0cm, 0cm, 0cm}, clip, width=\linewidth]{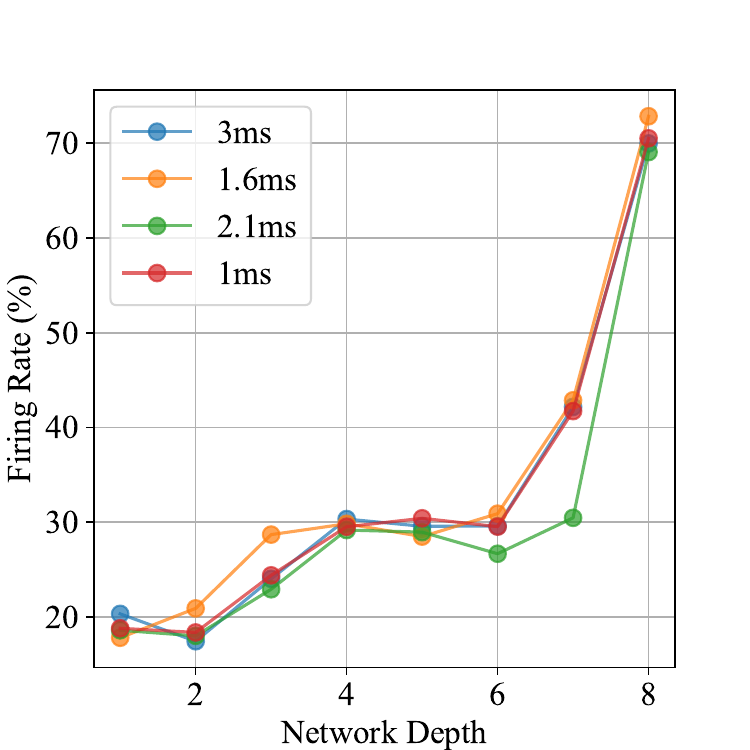}
            \caption{Fixed Time Window.}
        \end{subfigure} &
        \begin{subfigure}{0.5\linewidth}
            \centering
            \includegraphics[trim={0cm, 0cm, 0cm, 0cm}, clip, width=\linewidth]{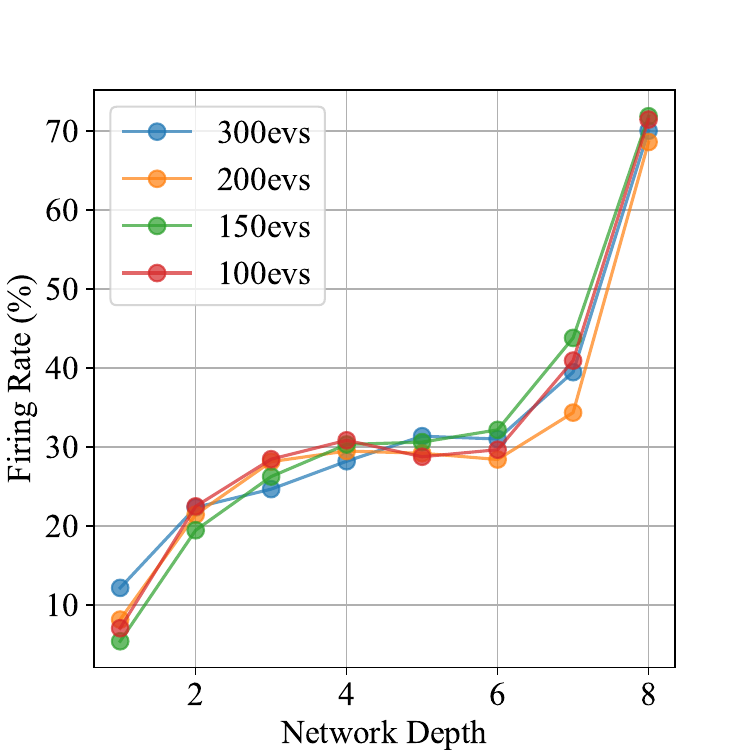}
            \caption{Dynamic Time Window.}
        \end{subfigure} \\
    \end{tabular}
    \caption{The firing rates of the trained \ac{snn} at different network depths with different slicing methods and time windows.}
    \label{tab:firing_rates}
\end{figure}

\paragraph{Firing Rates:}
\label{sec:firing_rate}
In \cref{tab:firing_rates}, we provide insights into the firing rates of the \ac{snn} at different network depths.  Results show the dynamic time window of events has considerably lower firing rate (10\% instead of 20\%) in the first layer. 

\subsubsection{Temporal Weighted-Sum Filter}
\label{sec:exp-ablation-model}

The temporal weighted-sum filter plays a crucial role in enhancing the performance of our model. In \cref{table:smoothing-filter}, we present an evaluation of the impact of two key parameters, namely $\tau_{mem}$ and $\tau_{syn}$, on the overall effectiveness of the filter. Notably, the  without the filter yields an error of 24.46 pixels ($\pm 3.17$). Adjusting the values of $\tau_{mem}$ and $\tau_{syn}$ allows for a tailored optimization of the filter's performance. 

% \begin{table}[h]
% \centering
% \caption{The performance of Retina for methods of spike to bounding box conversion.}
% \label{table:regression-snn}
% \small
% \begin{tabular}{c|c|c|c|c}
% \toprule 
% Filter & Spike Count & $V_{me}$ & Error (px)  $\downarrow$  & IoU (\%)  $\downarrow$  \\ 
%  \midrule
%   &   & \checkmark  & 25.71 ($\pm$ 3.17) & 2.01  \\
%   & \checkmark  &   & 24.46 ($\pm$ 1.96) & 2.72 \\ 
%  \checkmark & \checkmark  & &   \textbf{3.24 ($\pm$ 0.79)} & \textbf{41.42} \\
% \bottomrule
% \end{tabular}
% \end{table}

\begin{table}[h]
\centering
\caption{The effect of the components of the temporal filter.}
\label{table:smoothing-filter}
\small
\begin{tabular}{c|c|c|c|c}
\toprule 
Figure & $\tau_{mem}$ & $\tau_{syn}$ & Kernel Size & Error (px)  $\downarrow$  \\ 
 \midrule  
- & \multicolumn{3}{c|}{Not Used} & 24.46 ($\pm 3.17$) \\ \midrule
B &  5  & 1  & 20 & 21.70 ($\pm 4.54$) \\
C & 1  & 5  & 20 & 8.72 ($\pm 4.60$)  \\
D & 5  & 5  & 20 &  \textbf{3.24 ($\pm$ 0.79)} \\
E & 10  & 10  & 20 & 3.52 ($\pm$ 0.89)  \\  
\bottomrule
\end{tabular}

\includegraphics[trim={0cm, 0cm, 0cm, 0cm}, clip, width=\linewidth]{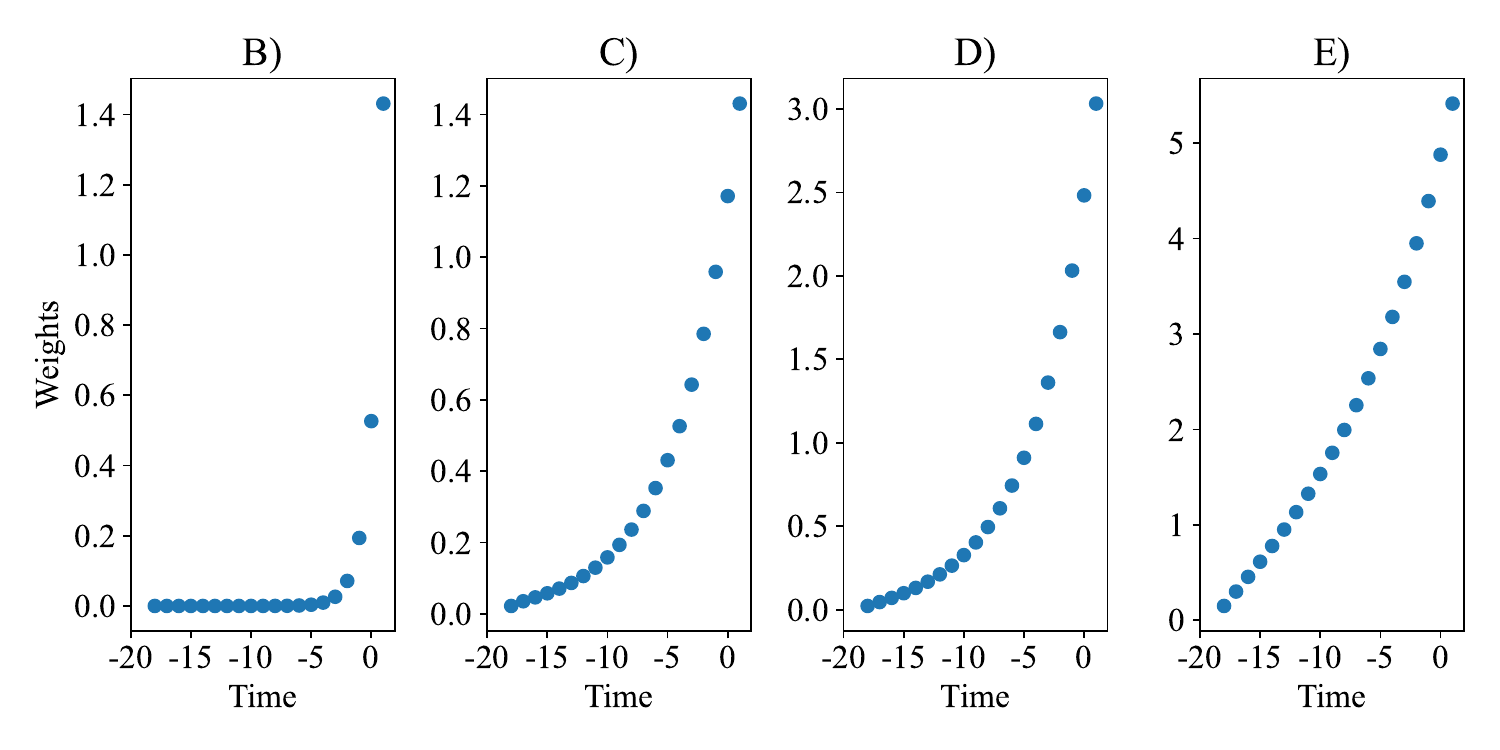} 
    %          \includegraphics[trim={0cm, 0cm, 1cm, 1cm}, clip,width=0.45\linewidth]{images/filter/5_1.png}  \\
    %          A) & B) \\
    %           \includegraphics[trim={0cm, 0cm, 1cm, 1cm}, clip,width=0.45\linewidth]{images/filter/1_5.png}  & 
    %          \includegraphics[trim={0cm, 0cm, 1cm, 1cm}, clip,width=0.45\linewidth]{images/filter/10_10.png} \\ 
    %          C) & D) \\
    %          \bottomrule
    % \end{tabular} 
    
\end{table}

\subsubsection{Architecture Components}
\label{sec:exp-ablation-model}

We conducted additional experiments on variants of the Retina model, specifically one without neuron state resets ("Retina w/o resets") during training and another trained with 3ET's loss function ("Retina w/o box"). The latter variant shares the same network architecture as Retina, except for the output layer, which directly predicts pupil coordinates.

\begin{table}[h]
\centering
\caption{The performance of our model on the validation set.}
\label{table:archi-centroid-error}
\small
\begin{tabular}{c|c}
\toprule
Method &  Error (px) $\downarrow$  \\
 \midrule
Retina  w/o  ${box}$ &  5.89 ($\pm$ 1.71) \\
Retina  w/o  ${reset}$ & 7.99 ($\pm$ 5.79)\\
Retina  &   \textbf{3.24 ($\pm$ 0.79)} \\   
\bottomrule
\end{tabular}
\end{table}

\cref{table:archi-centroid-error} shows the importance of resetting the neuron states during training and the efficacy of predicting bounding boxes instead of single pixels coordinates.

\subsubsection{Latency \& Power Consumption}
\label{sec:exp-latency-power}
In this section, we present a comprehensive analysis of the power consumption, energy and latency metrics on the \ac{speck} platform for the two time windows of events, namely the Dynamic Window (300 events) and the Fixed Window ($3 \text{ms}$). We injected the events bins at sufficiently distant timestamps ($100ms$) in order to isolate the response of the chip. \cref{table:power-latency} provides an overview of the power consumption across various components contributing to the overall power profile in the \ac{speck} namely:  Vdd, Vda, (referring to the digital and analog steps on the \ac{dvs}), Logic (the power usage of the \ac{snn}), RAM, and I/O. The Dynamic Window can be used to stabilize the power consumption in case higher rates of events are expected, however in our test the Fixed Window was able to utilize the resources available more efficiently. We computed the latency by calculating the difference between the timestamp for any given bin from the input timestamp of the events to the generation of spikes (on average 1'700) for a prediction. 

\begin{table}[h]
\centering
\caption{The average and peak latency (L - $ms$),  power consumption (P - mW) and energy (E - $mJ$) on \ac{speck} for the validation dataset using two time windows of events.}
\label{table:power-latency}
\small
\begin{tabular}{c|c|c|c|c} 
\toprule
Device & Channel & Unit & \text{Dynamic} & \textbf{Fixed}  \\  \midrule
\multirow{4}{*}{\ac{dvs}} & \multirow{2}{*}{Vdd} & \multirow{10}{*}{$mW$}& \multicolumn{2}{c}{0.03} \\ 
       & && \multicolumn{2}{c}{[0.01-0.05]} \\ \cmidrule{4-5} \cmidrule{2-2} 
 & \multirow{2}{*}{Vda} &   &\multicolumn{2}{c}{0.6} \\ 
       & && \multicolumn{2}{c}{[0.58-0.63]} \\ \cmidrule{4-5} \cmidrule{1-2}
       
\multirow{7}{*}{Processor} & \multirow{2}{*}{Logic} &  &2.43 & \textbf{1.26} \\ 
       &&& [0.28-14.34] &  [0.28-15.08] \\ \cmidrule{4-5} \cmidrule{2-2} 
 & \multirow{2}{*}{RAM} &   &1.64 & \textbf{0.90}\\ 
 &&& [0.01-9.44] & [0.03-8.95] \\ \cmidrule{4-5} \cmidrule{2-2} 
 & \multirow{2}{*}{I/O} &   & \multicolumn{2}{c}{0.10} \\ 
 &  &&\multicolumn{2}{c}{[0.08-0.23]} \\ \midrule  \midrule
\multirow{4}{*}{Total} & \multirow{4}{*}{End-to-End}  & $mW$ &  4.80 & \textbf{2.89} \\ \cmidrule{3-5} 
 &  & $ms$ & 8.01 & \textbf{5.57} \\  \cmidrule{3-5} 
 & & $mJ$ & 38.40 & \textbf{16.10} \\ 
\bottomrule
\end{tabular}
\end{table}

\subsection{Benchmark Comparison} 

 We quantified the efficacy of our system using the Centroid Error metric. The validation protocol employed a leave-two out participant scheme for Ini-30. \cref{table:centroid-error} presents the centroid error results for a \ac{dvs} resolution of 64x64x2. The results of 3ET on the synthetic dataset are different compared to those reported in the original manuscript because we modified the data pipeline to transmit only 1-bit event data. 
 
\begin{table}[h]
\centering
\caption{The performance of our model on the validation set.}
\label{table:centroid-error}
\small
\begin{tabular}{c|c|c}
\toprule
Dataset & 3ET \cite{chen20233et} & Retina \\
 \midrule
Ini-30 & 4.48 ($\pm$ 1.94) & \textbf{3.24 ($\pm$ 0.79)}\\  
Synthetic (LPW) \cite{chen20233et} &  \textbf{5.33 ($\pm$ 1.59)} & 6.46 ($\pm$ 2.49)\\  
\bottomrule
\end{tabular}
\end{table}

In \cref{table:mac-params}, we present a comprehensive comparison between our proposed model and the state-of-the-art 3ET \cite{chen20233et}, focusing on both the number of parameters and the volume of \ac{mac} operations. Notably, our model demonstrates a remarkable reduction in complexity, underscoring its efficiency.

\begin{table}[h]
\centering
\caption{A breakdown of the network complexity.}
\label{table:mac-params}
\small
\begin{tabular}{c|c|c}
\toprule
Method &  MAC Operations $\downarrow$ & Parameters $\downarrow$ \\
\midrule 
3ET \cite{chen20233et}  & 107M & 418k \\
Retina &  3.03M & 63k \\  
\bottomrule
\end{tabular}
\end{table}

\section{Discussion and Future Work}
As shown in \cref{table:archi-centroid-error}, Retina is less precise when trained without the continuous resetting of neuron states. The ongoing reset of neuron states may introduce potential disruptions in continuous tracking on a neuromorphic chip. Exploring strategies to mitigate this challenge and optimize model performance under those additional hardware constraints is a promising avenue for future research.

\section{Conclusion}
Our work describes an energy-efficient (5$mW$, end-to-end), low latency (6$ms$, end-to-end), and accurate (3-4\si{px}) neuromorphic approach for eye tracking, leveraging the strengths of the neuromorphic form of both sensor and processor, and a truly lightweight and deployable spiking neural network model. The presented model demonstrates better than baseline precision with significantly reduced computational complexity. 
%The introduced event-based eye-tracking dataset also represents a significant step forward in the realm of real-world ultra-low power eye-tracking technology. 
Finally, we hope the introduced event-based eye-tracking dataset Ini-30 could promote further exploration in the realm of real-world ultra-low power wearable eye-tracking technology. 
{
    \small
    \bibliographystyle{ieeenat_fullname}
    \bibliography{main}
}

% WARNING: do not forget to delete the supplementary pages from your submission 
%\input{sec/X_suppl}

\end{document}